\documentclass{article}

\pdfoutput=1

    \PassOptionsToPackage{numbers, compress}{natbib}
    \usepackage[preprint]{neurips_2020}

\usepackage[utf8]{inputenc} % allow utf-8 input
\usepackage[T1]{fontenc}    % use 8-bit T1 fonts
\usepackage{hyperref}       % hyperlinks
\usepackage{url}            % simple URL typesetting
\usepackage{booktabs}       % professional-quality tables
\usepackage{amsfonts}       % blackboard math symbols
\usepackage{nicefrac}       % compact symbols for 1/2, etc.
\usepackage{microtype}      % microtypography
\usepackage{graphicx}
\usepackage{color}

\usepackage[ruled,vlined]{algorithm2e}

\newcommand{\etal}{\textit{et al.}}

\usepackage{amsmath,amsfonts,bm}

\def\eqref#1{equation~\ref{#1}}
\def\1{\bm{1}}

\def\eps{{\epsilon}}

\def\rmQ{{\mathbf{Q}}}

\def\vl{{\bm{l}}}

\def\vq{{\bm{q}}}
\def\vr{{\bm{r}}}

\def\vx{{\bm{x}}}
\def\vy{{\bm{y}}}

\DeclareMathAlphabet{\mathsfit}{\encodingdefault}{\sfdefault}{m}{sl}
\SetMathAlphabet{\mathsfit}{bold}{\encodingdefault}{\sfdefault}{bx}{n}

\def\data{\vx}

\def\perturb{\bm{\eta}}

\def\gen{\bm{G}}

\def\eps{\bm{\epsilon}}

\def\rTarget{\vr}

\def\nqueries{\bm{B}}
\def\queries{\rmQ}

\def\totalQueries{\nqueries_0}

\def\dataset{\mathcal{D}}
\def\leakDataset{\dataset_s}
\def\perturbSet{\mathcal{P}}
\def\component{\vq}
\def\componentSet{\mathcal{Q}}

\def\norm{p}
\def\classifier{\mathit{h}}
\def\lbl{\vy}

\def\threshold{\tau}

\title{Protecting Against Image Translation Deepfakes \\ by Leaking Universal Perturbations \\ from Black-Box Neural Networks}

\author{%
  Nataniel Ruiz \\
  Department of Computer Science\\
  Boston University\\
  \texttt{nruiz9@bu.edu} \\
  \And
  Sarah Adel Bargal \\
  Department of Computer Science\\
  Boston University\\
  \texttt{sbargal@bu.edu}  \\
  \AND
  Stan Sclaroff \\
  Department of Computer Science\\
  Boston University\\
  \texttt{sclaroff@bu.edu} \\
}

\begin{document}

\maketitle

\begin{abstract}

In this work, we develop efficient disruptions of black-box image translation deepfake generation systems. We are the first to demonstrate black-box deepfake generation disruption by presenting image translation formulations of attacks initially proposed for classification models. Nevertheless, a naive adaptation of classification black-box attacks results in a prohibitive number of queries for image translation systems in the real-world. We present a frustratingly simple yet highly effective algorithm \textit{Leaking Universal Perturbations (LUP)}, that significantly reduces the number of queries needed to attack an image. LUP consists of two phases: (1) a short \textit{leaking phase} where we attack the network using traditional black-box attacks and gather information on successful attacks on a small dataset and (2) and an \textit{exploitation phase} where we leverage said information to subsequently attack the network with improved efficiency. Our attack reduces the total number of queries necessary to attack GANimation and StarGAN by \textit{30\%}.

\end{abstract}

\section{Introduction}

The popular term ``deepfake'' originally referred to the deep learning technique of transforming the face of a person to reflect a different identity. Original techniques used a shared encoder and distinct decoders for separate identities~\cite{guera2018deepfake}. Contemporaneous methods using more elaborate combinations of deep learning with computer graphics techniques achieved impressive results~\cite{kim2018deep, thies2016face2face, thies2018headon, Usman_2019_ICCV}.

Since then the term ``deepfake'' has been adopted in a broader context and can be used to refer to any altered media of someone's likeness. Recently there have been remarkable advances in face modification algorithms. Some algorithms only need a single image and can create modified versions of that person under different poses, expressions, lighting and other attribute changes~\cite{choi2018stargan, pumarola2018ganimation, choi2019stargan}. The most advanced algorithms can create puppeteering videos using as few as one image~\cite{zakharov2019few, tewari2020stylerig}. This few-shot deepfake technology based on image translation networks has gained popularity in the mainstream with apps such as FaceApp~\footnote{\url{https://www.faceapp.com}} that allow for transformation of images such as putting a smile on someone's face and making them appear older or younger, among other interventions.

These technologies can be used in malicious ways to produce undesirable content of someone without their consent. This concern has already materialized in several ways, including creating non-consensual pornographic footage and producing videos with fake political speeches.

Attempts to detect manipulated media are underway and there is an ``arms race'' between detecting deepfakes~\cite{Rossler_2019_ICCV,8683164,li2019exposing,wangfakespotter,wang2019cnngenerated} and evasion of deepfake detection~\cite{neekhara2020adversarial,gandhi2020adversarial}. Instead of detecting deepfakes after the fact, Ruiz \etal~\cite{ruiz2020disrupting} recently proposed using \textit{white-box adversarial attacks} to protect an image from modification by image translation networks. While this work assumes that the adversary has access to the model's structure, weights and gradients, in a real scenario, these might not be accessible. In this work, we focus on the \textit{black-box scenario} where model parameters are unknown.

An image translation-based online deepfake generation service usually allows for API queries where a user sends an image and receives the translated output. This is an instance of the image translation black-box threat model, where the structure, weights and gradients of the black-box are unknown, but the output images are available to the adversary. In this work we demonstrate attacks on image translation models under this threat model and show the vulnerability of several popular image translation networks. Specifically, we are the first to explore black-box adversarial attacks on image translation systems with an application of disrupting deepfake generation.

In our work, we reformulate black-box attacks for the image translation scenario and demonstrate their effectiveness in preventing deepfake generation. However, the number of queries of such black-box attacks might be prohibitive in a real-world scenario where an adversary might detect an attempted attack or the query budget might run out. We present a simple, yet highly effective, algorithm that we call \textit{Leaking Universal Perturbations (LUP)} that sharply decreases the average number of queries required to generate attacks. LUP is composed of two phases, a short \textit{leaking phase} during which the network is attacked using a traditional black-box attack on a small dataset of images. And an \textit{exploitation phase}, where the algorithm leverages the information obtained during the \textit{leaking phase} to subsequently attack the network with improved efficiency.

During the LUP \textit{leaking phase} we attack a set of images using any black-box attack method. Once these perturbations have been generated, PCA components are extracted from them. During the \textit{exploitation phase} we use a modified Simple Black-Box Attack (SimBA)~\cite{guo2019simple} where the \textit{leaking phase} PCA components are used as attack vectors. If the image has not yet been successfully attacked, and the loss saturates during the \textit{exploitation phase} the algorithm reverts to the default image basis to finish the attack. We are able to reduce the average number of necessary queries to attack an image during the \textit{exploitation phase} by around 30\% compared to state-of-the-art methods such as SimBA on multiple image translation networks.

We summarize our contributions as follows:

\begin{itemize}
    \item We are the first to successfully attack black-box image translation systems, with an application to disrupting generation of deepfake images. This is a first step in protecting photographs of people from being modified by facial manipulation systems in the real world.
    \item We present a novel method called \textit{leaking universal perturbations (LUP)}, that vastly improves the efficiency of black-box attacks by exploiting information gathered during initial attacks performed in a \textit{leaking phase}. This allows the attack to scale more efficiently compared to other state-of-the-art methods.
\end{itemize}

\section{Related Work}

\paragraph{White-box Attacks on Classifiers}

Different threat models for adversarial attacks have been defined for the image classification scenario. They are defined by the amount of information that the adversary has regarding the target model. Under a white-box threat model in the classification scenario, the structure and weights of the classifier $h$ are available to the adversary. This means that the classifier can be run locally on the adversaries' infrastructure, and gradients can be computed. Under this threat model Szegedy \etal~\cite{szegedy2013intriguing} demonstrated the existence of adversarial examples for deep neural network classifiers. Since then, there has been a large amount of work on attacking models under this setting by performing gradient descent on the defined classification loss $l$ or optimization methods using the gradient information~\cite{szegedy2013intriguing,explaining_adv,moosavi2016deepfool,papernot2016limitations,carlini2017towards,nguyen2015deep, moosavi2017universal,kurakin2016adversarial,madry2018towards}.

\paragraph{Black-box Attacks on Classifiers}

In a real-world scenario the adversary might not have access to either the structure or the weights of the classifier $h$. Instead, she might have access to an API which allows queries to the model. The adversary might then have either access to the probability outputs, or uniquely to the classification decisions of the model. The goal is to attack the model while minimizing the number of queries as well as the magnitude of the attack under a suitable norm. 

There is extensive work on black-box attacks on classification deep networks. One approach is to train a surrogate network and transfer white-box attacks generated using the surrogate network to the target network~\cite{papernot2017practical,liu2016delving}. Another effective approach is to estimate the gradients using finite-differences, Monte Carlo sampling methods or other techniques and subsequently perform gradient descent~\cite{chen2017zoo,ilyas2018black,IEM2018PriorCB,Cheng2019QueryEfficientHB,tu2019autozoom}. Another class of approaches are local-search approaches that attack the network by probing the black-box without any gradient estimation~\cite{narodytska2016simple, guo2019simple}. In contrast to these methods, our LUP attack learns to perform a more efficient attack by collecting information on initial black-box attacks.
In essence, LUP learns transferable attack components that can be used to efficiently query the model in a local-search manner in subsequent attacks.

\paragraph{Image Translation Adversarial Attacks}

Image translation networks have recently achieved impressive results in deepfake generation and face modification using few images (or one image) of an individual~\cite{choi2018stargan, pumarola2018ganimation, choi2019stargan, zakharov2019few, tewari2020stylerig}. Some models allow for generation of video of a person saying things that they did not say, using a single image~\cite{zakharov2019few, tewari2020stylerig}. In general, most image translation models are trained using a GAN setup. Some are trained in a supervised manner~\cite{isola2017image, wang2018high, zakharov2019few}, while others are trained in an unsupervised manner~\cite{zhu2017unpaired, pumarola2018ganimation, choi2018stargan, choi2019stargan}.

There is previous work that demonstrates attacks on generative models and more specifically autoencoders~\cite{tabacof2016adversarial, kos2018adversarial}. There is also work that explores self-adversarial attacks during training of unsupervised GAN-based image translation networks~\cite{chu2017cyclegan,bashkirova2019adversarial}. Recently, there has been work that proposes adversarial attacks on image translation networks with the goal of disrupting the output~\cite{ruiz2020disrupting}. Disrupting generation of deepfakes images reveals itself to be an interesting application for these types of attacks. Ruiz \etal~\cite{ruiz2020disrupting} explore white-box attacks on image translation networks such as pix2pixHD~\cite{wang2018high} and CycleGAN~\cite{zhu2017unpaired}. They also show that their white-box attack allows for disruption of deepfake generation using StarGAN~\cite{choi2018stargan} and GANimation~\cite{pumarola2018ganimation}. Ruiz \etal~\cite{ruiz2020disrupting} explore white-box attacks on image translation models to disrupt deepfake generation, in contrast we explore black-box attacks.

\section{Method}

We first provide formulations for image translation attacks (Sec.~\ref{B}). We then present our formulations of black-box attacks for image translation (Sec.~\ref{BBIT}). We then propose our method of leaking universal perturbations (LUP), that obtains more efficient attacks than competing methods (Sec.~\ref{LUPS}).

\subsection{Background}
\label{B}

An adversarial example is an image with small additive changes, that can be imperceptible to a human being, and affect the output label of the image classification model. In general an adversarial attack, which creates adversarial examples, on an image classification model $h$ is defined by:
\begin{equation}
    \min_{\perturb}l_{\lbl}(\classifier(\data + \perturb))
    \text{, ~~~}  
    \text{subject to } \norm(\perturb) \leq \eps.
\end{equation}
Different distance norms $\norm$ have been proposed, and attacks usually use the $L_2$ or $L_\infty$ norms. $l_\lbl$ is a surrogate loss that measures the degree of certainty that the model will classify the input as class $\lbl$. This surrogate loss can be defined in different ways, depending on the output of the model.

To delve into black-box attacks on image translation models, it is helpful to first present formulations for the white-box scenario. Ruiz \etal~\cite{ruiz2020disrupting} formulate a targeted attack on an image translation generator $\gen$, with target $\rTarget$:
\begin{equation}
    \min_{\perturb}L(\gen(\data + \perturb), \rTarget)
    \text{, ~~~}  
    \text{subject to } \norm(\perturb) \leq \eps,
\end{equation}
where $\data$ is the input image, $\perturb$ is the generated perturbation, $\norm$ is a chosen norm, $\eps$ is the maximum attack magnitude and $L$ is the chosen image-level regression loss.

They also define an untargeted attack seeking to maximize the distortion of the output image with respect to the ground-truth non-attacked output. We call this a \textit{maximum distortion attack}.
\begin{equation}
    \max_{\perturb}L(\gen(\data + \perturb), \gen(\data)) \text{, ~~~}  
    \text{subject to } \norm(\perturb) \leq \eps.
\end{equation}

\subsection{Black-Box Attacks on Image Translation Models}
\label{BBIT}

In this section we propose formulations of image translation formulations of black-box attacks initially proposed for image classification. We reformulate two gradient estimation-based approaches, Natural Evolution Strategies~\cite{ilyas2018black} and Bandits-TD~\cite{IEM2018PriorCB} and a local search-based attack SimBA~\cite{guo2019simple}. We do so by replacing the typical classification loss $l$ by an image-level regression loss $L$, that measures the distance between the target image $\rTarget$ and the translated adversarial example $\gen(\data + \perturb)$.

\paragraph{Gradient Estimation-based Attacks}
Under a black-box threat model, the weights and structure of the model are unknown. Thus, attacking the model using gradient-descent is not directly feasible. A solution explored in the literature on attacks on classifiers is to estimate the gradient $\nabla l(\data)$ of the classification loss $l$ around point $\data$ by using Monte Carlo sampling methods.

We reformulate both the Natural Evolution Strategies (NES)~\cite{ilyas2018black} and Bandits-TD~\cite{IEM2018PriorCB} attacks for the image translation scenario. We show that they are able to produce effective attacks on popular image translation networks and are able to disrupt deepfake generation.

Our formulation of image translation NES (IT-NES) gradient estimate for the image-level regression loss $L$ using generator $\gen$, with $n$ queries is:
\begin{equation}
\nabla\mathbb{E}[L(\gen(\data), \rTarget)] \approx \frac{1}{\sigma n}\sum_{i=1}^n \delta_i
L(\gen(\data + \sigma\delta_i), \rTarget),
\end{equation}
where $\rTarget$ is the target image, $\sigma$ is the variance of the Gaussian search distribution and using antithetic sampling we have $\delta_i \sim \mathcal{N}(0, I)$ for $i \in \{1, \ldots, \frac{n}{2}\}$ and set $\delta_j = -\delta_{n-j+1}$ for $j \in \{(\frac{n}{2}+1), \ldots, n\}$. The adversarial example is then updated using the estimated gradient 
\begin{equation}
\data_{t+1} = \data_{t} - \eps \nabla\mathbb{E}[L(\gen(\data), \rTarget)].
\end{equation}

Bandits-TD introduces both a time dependent prior capturing the correlation of gradients across time steps and a data dependent prior capturing the spatial correlation of gradients in an image. Bandits-TD uses the antithetic NES gradient estimation method with $n=2$.

We reformulate the Bandits-TD black-box attack for image translation in similar fashion to IT-NES.  We call this formulation image translation Bandits-TD (IT-Bandits-TD). We replace the classification criterion $l_\lbl(\data)$ for image $\data$ and label $\lbl$ by the image-level regression criterion $L_\rTarget = L(\data, \rTarget)$, where $\rTarget$ is the target image. We perform gradient descent for our targeted attack.

\paragraph{Local Search-based Attack}

We reformulate the Simple Black-box Attack (SimBA) proposed in \cite{guo2019simple} for the image translation scenario. First, SimBA generates an orthonormal image basis of vectors $\component \in \Gamma$, and $\Gamma$ is the set of orthonormal candidate vectors. A natural basis is composed of images with $0$ everywhere except at $(i,j)$ for all $i \in \{1,...,d\}$ and $j \in \{1,...,d\}$, where $d$ is the image size.

In the classification scenario, with a classifier $h$ and a classification loss $l_\lbl(h(\data))$, SimBA iterates over all candidate vectors $\component \in \Gamma$. At time step $t$, SimBA determines whether $\data + \component$ increases the loss. If yes, the candidate vector is added to the adversarial example, if not then the opposite direction is sampled $\data - \component$ to the same effect. If both directions do not increase the loss then the candidate vector is skipped. The algorithm halts whenever the classifier incorrectly classifies the perturbed image $h(\tilde{\data}) \neq \lbl$. Again, we reformulate SimBA for image translation by replacing the classification loss $l_\lbl$ by the regression loss $L_\rTarget$, and call this method image translation SimBA (IT-SimBA).

\subsection{Leaking Universal Perturbations (LUP)}
\label{LUPS}

In the black-box adversarial attack setting, we are given a budget of black-box queries for each image we would like to attack. In this setting, we have the same number of maximum allowed queries for all images in the dataset. That is, for each image $\data$ we want to solve the optimization problem 
\begin{equation}
    \min_{\perturb}L(\gen(\data + \perturb), \rTarget) \text{, ~~~}  
    \text{subject to } \norm(\perturb) \leq \eps, \queries \leq \nqueries,
\end{equation}
where $\gen$ is the generator of the image translation system, $\perturb$ is the perturbation, $\queries$ is the number of queries used and $\nqueries$ is the maximum number of queries allowed for a single image.

An adversary would benefit from reducing the \textit{total number of queries} required to attack a given dataset $\totalQueries$. Our proposed algorithm seeks to reduce $\totalQueries$ by, first, leaking elements of a universal perturbation from a small auxiliary dataset and then exploiting these transferable components on the images in the larger test dataset. It has two phases (1) the \textit{leaking phase}, where we attack the model using a traditional attack and gather information on successful attacks on a small auxiliary dataset (2) the \textit{exploitation phase} where we attack the model using the leaked information from the first phase on the larger test set. This allows us to sharply reduce the number of amortized queries needed.

\begin{minipage}{0.52\textwidth}
\begin{algorithm}[H]
\SetAlgoLined
 \KwIn{image $\data$, generator $G$, leaked PCA components $\componentSet$, image basis $\Gamma$, threshold $\threshold$, max. queries $\nqueries$, step size $\xi$, max. saturating loss iterations $n_{\text{sat}}$, target image $\rTarget$}
 \KwOut{perturbation $\perturb$}
 $\perturb = 0$; \ $\queries = 0$; \ $i = 0$\;
 phase two = False\;
 $\vl = L(\gen(x), \rTarget)$\;
 \While{$\vl < \threshold$ and $\queries \leq \nqueries$} {
  \eIf{phase two}
  {
    pick $\component \in \Gamma$ randomly w/o replacement\;
  }
  {
    pick $\component \in \componentSet$ in order\;
  }
  \For{$\alpha \in \{\xi, -\xi\}$} {
  $\vl' = L(\gen(\data + \perturb + \alpha \component), \rTarget)$\;
  $\queries = \queries + 1$\;
  \eIf{$\vl' < \vl$}{
  $\perturb = \perturb + \alpha \component$\;
  $\vl = \vl'$\;
  $i = 0$\;
  \textbf{break}\;
  }
  {
  $i = i + 1$
  }
  }
  \If{$i \geq n_{\text{sat}} - 1$}
  {
  phase two = True\;
  }
  }
 \Return{$\perturb$}
 \caption{LUP Exploitation Phase} 
 \label{alg1}
\end{algorithm}
%\vspace{-10pt}
\end{minipage}
\hfill
\begin{minipage}{0.49\textwidth}
\begin{algorithm}[H]
\SetAlgoLined
 \KwIn{
 leaking dataset $\leakDataset$, generator $G$, image basis $\Gamma$, threshold $\threshold$, max. queries $\nqueries$, step size $\xi$, target image $\rTarget$}
 \KwOut{set of leaked PCA components $\componentSet$}
 $H = \{\}$\;
 \For{$\data \in \dataset$} {
  $\perturb = \text{IT-SimBA}(\data,\gen,\Gamma,\threshold,\nqueries,\xi,\rTarget)$\;
  $H = H \cup \{\perturb\}$\;
 }
 run PCA on $H$ and obtain components $\componentSet$\;
 \Return{$\componentSet$}
 \caption{LUP Leaking Phase}
 \label{alg2}
\end{algorithm}

\vspace{0pt}

\begin{algorithm}[H]
\SetAlgoLined
 \KwIn{image $\data$, generator $G$, image basis $\Gamma$, threshold $\threshold$, max. queries $\nqueries$, step size $\xi$, target image $\rTarget$}
 \KwOut{perturbation $\perturb$}
 $\perturb = 0$; \ $\queries = 0$; \ $i = 0$\;
 $\vl = L(\gen(x), \rTarget)$\;
 \While{$\vl < \threshold$ and $\queries \leq \nqueries$} {
  pick $\component \in \Gamma$ randomly w/o replacement\;
  \For{$\alpha \in \{\xi, -\xi\}$} {
  $\vl' = L(\gen(\data + \perturb + \alpha \component), \rTarget)$\;
  $\queries = \queries + 1$\;
  \If{$\vl' < \vl$}{
  $\perturb = \perturb + \alpha \component$\;
  $\vl = \vl'$\;
  \textbf{break}\;
  }
  }
  }
 \Return{$\perturb$}
 \caption{IT-SimBA: \textit{SimBA~\cite{guo2019simple} using an image-level regression loss $L$ for attacking image translation models}}
\end{algorithm}
%\vspace{-10pt}
\end{minipage}

\paragraph{Algorithm}
Our algorithm has two phases, the \textit{leaking phase} and the \textit{exploitation phase}. During the \textit{leaking phase}, it performs a traditional black-box attack on a separate dataset $\leakDataset$ consisting of $N_s$ images, drawn from the same distribution as our test dataset $\dataset$. We extract principal components from these perturbations using principal component analysis (PCA). During the \textit{exploitation phase} we use the principal components to improve the efficiency of our black-box attacks on the test dataset $\dataset$. We accomplish this by querying the black-box using the leaked principal components using a modified IT-SimBA. We achieve strong attacks using fewer queries. We now describe the two phases in detail, and provide pseudocode in Algorithms~\ref{alg1},~\ref{alg2}.

\paragraph{Leaking Phase}
During the \textit{leaking phase} we apply a black-box attack on a leaking dataset $\leakDataset$. We attack all images $\data \in \leakDataset$ until we achieve successful attacks ($L(\gen(\data + \perturb), \rTarget) < \threshold$, where $\threshold$ is the success threshold) or until we use a maximum number of $\queries$ queries. We create a set $\perturbSet$ of generated perturbations $\perturb$. Our framework is general and any attack or combination of attacks can be used for the leaking phase. We use PCA on perturbations $\perturb \in \perturbSet$ and extract principal components $\component \in \componentSet$.

\paragraph{Exploitation Phase}
Our \textit{exploitation phase} consists of using a modified IT-SimBA using $\component \in \componentSet$ as candidate vectors. Since $\componentSet$ is not necessarily a basis of the image space (because $N_s < d^2$), and even though the initial iterations of the attack very rapidly decrease the loss, the attack might saturate. We switch to a full basis in image space $\Gamma$ after a number of iterations $n_\text{sat}$ of saturating loss. The resulting attacks achieve strong results using substantially fewer queries $\queries$.

\section{Experiments}
\label{4}

In this section we first introduce the datasets and architectures used to demonstrate the efficacy of our LUP method (Sec. \ref{4_1}). When then introduce our experimental setup and implementation details. Next, we present results on two image translation-based deepfake generation architectures GANimation~\cite{pumarola2018ganimation} and StarGAN~\cite{choi2018stargan}. We use IT-SimBA for the \textit{leaking phase} in all experiments.

\subsection{Experimental Setup}
\label{4_1}

\begin{table}[t]
  \caption{Attack comparison on GANimation with threshold $\threshold = 0.005$ using 3 expressions (left) and on StarGAN for threshold $\threshold = 0.05$ using 5 attributes (right). We show the mean number of queries per image, as well as the average norm of the perturbation and the success rate percentage.
  \label{table:attack_comparison}}
\resizebox{0.49\textwidth}{!}{
  \begin{tabular}{cccc}
    \multicolumn{4}{c}{\large \hspace{3ex} \textbf{GANimation}} \\
    \toprule
    Attack & Avg. Queries & Avg. Norm & Success Rate \\
    \midrule
    IT-NES & 598 & 1.82 & 98.8\% \\
    IT-Bandits-TD & 855 & 4.38 & 96.3\%  \\
    IT-SimBA & 551 & 4.87 & 97.9\% \\
    LUP & \textbf{393} & 3.07 & 98.6\% \\
    \bottomrule
  \end{tabular}
}
\resizebox{0.49\textwidth}{!}{
  \begin{tabular}{cccc}
    \multicolumn{4}{c}{\large \hspace{3ex} \textbf{StarGAN}} \\
    \toprule
    Attack & Avg. Queries & Avg. Norm & Success Rate \\
    \midrule
    IT-NES & 1,001 & 2.90 & 99.8\% \\
    IT-Bandits-TD & 4,901 & 4.99 & 52.2\%  \\
    IT-SimBA & 444 & 5.93 & 100\% \\
    LUP & \textbf{313} & 6.36 & 100\% \\
    \bottomrule
  \end{tabular}
}
   \vspace*{0pt}
\end{table}

\begin{figure*}[t]
    \centering
    \includegraphics[width=0.4\textwidth]{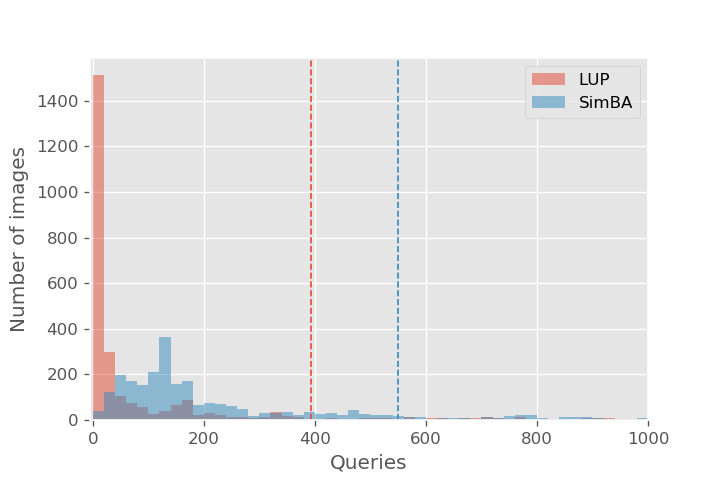}
    \includegraphics[width=0.4\textwidth]{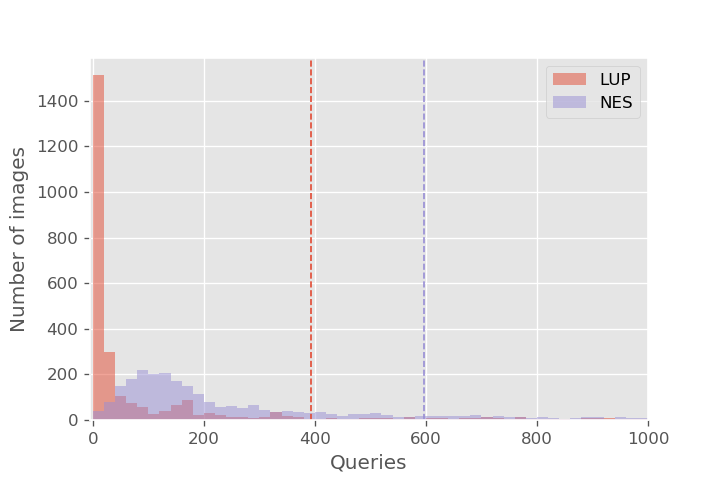}
    \vspace{-7pt}
    \caption{
    Histogram of queries required for successful attacks on GANimation (3,000 attacks, three expressions), using success threshold $\threshold=0.005$. Vertical lines show mean queries for each method.
    \label{fig:ganimation_0005_hist}}
\end{figure*}

\begin{figure}[t]
\includegraphics[clip,width=0.5\columnwidth]{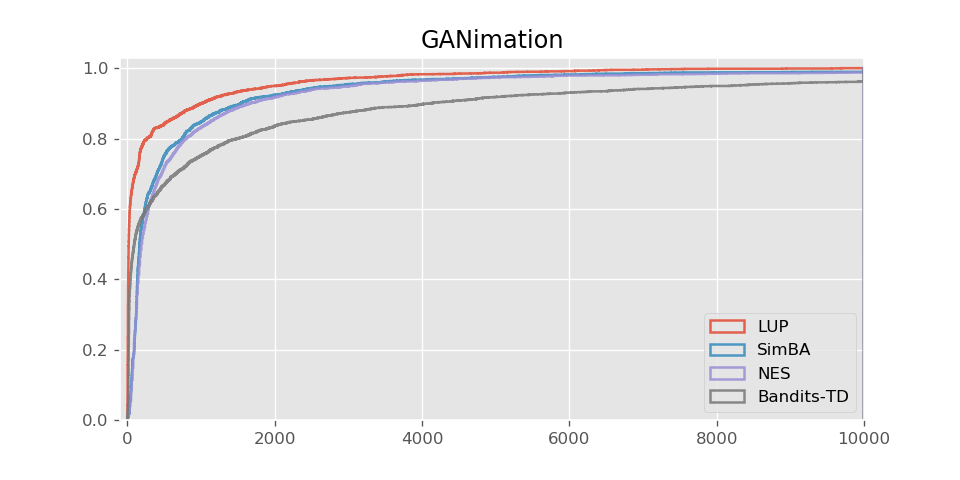}
\includegraphics[clip,width=0.5\columnwidth]{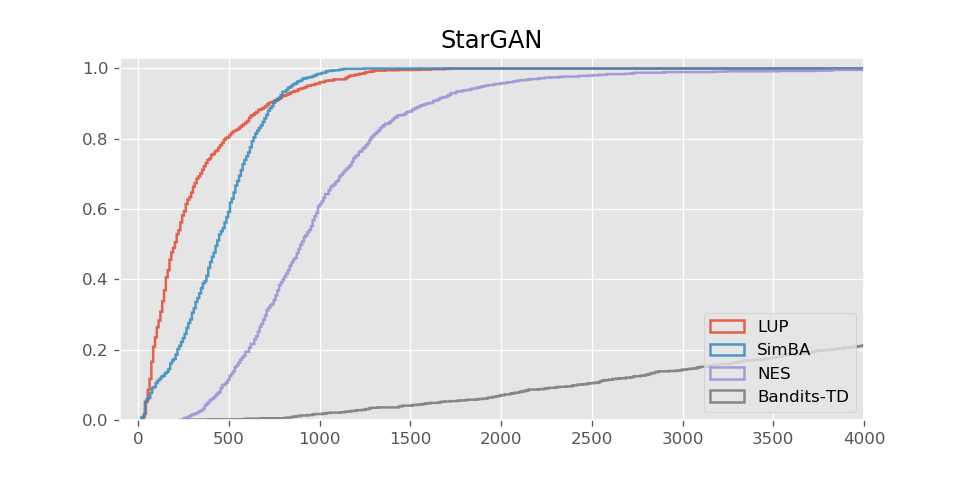}
\vspace{-20pt}
\caption[]{Success rate by number of queries on 3,000 attacks on GANimation for three expressions (left) and 1,000 attacks on StarGAN for five attribute classes (right). We use success thresholds $\threshold=0.005$ and $\threshold=0.05$ respectively. Vertical lines represent mean queries for each method.
\label{fig:algos_cumulative_0005} \vspace{-10pt}}
\end{figure}

\begin{figure*}[t]
    \centering
    \includegraphics[width=0.4\textwidth]{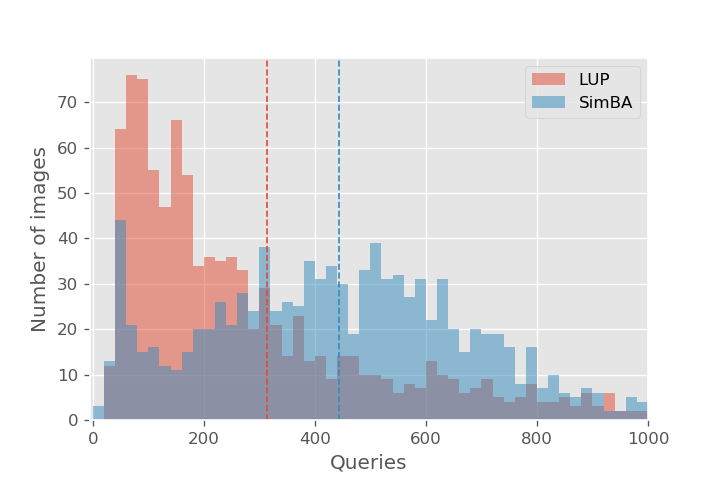}
    \includegraphics[width=0.4\textwidth]{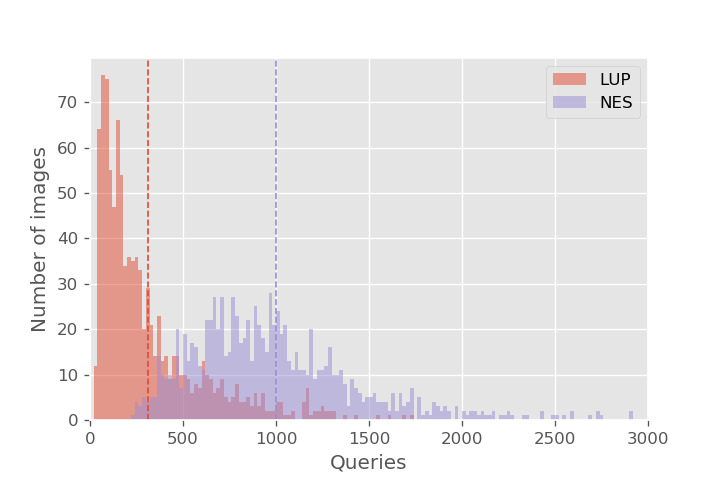}
    \vspace{-7pt}
    \caption{Histogram of queries required for a successful attack on StarGAN of 1,000 attacks (five classes). We use a success threshold $\threshold=0.05$. Vertical lines represent mean queries for each method.
    \label{fig:stargan_0050_hist}}
    \vspace{-10pt}
\end{figure*}

\begin{figure*}[t]
    \centering
    \includegraphics[width=0.8\textwidth]{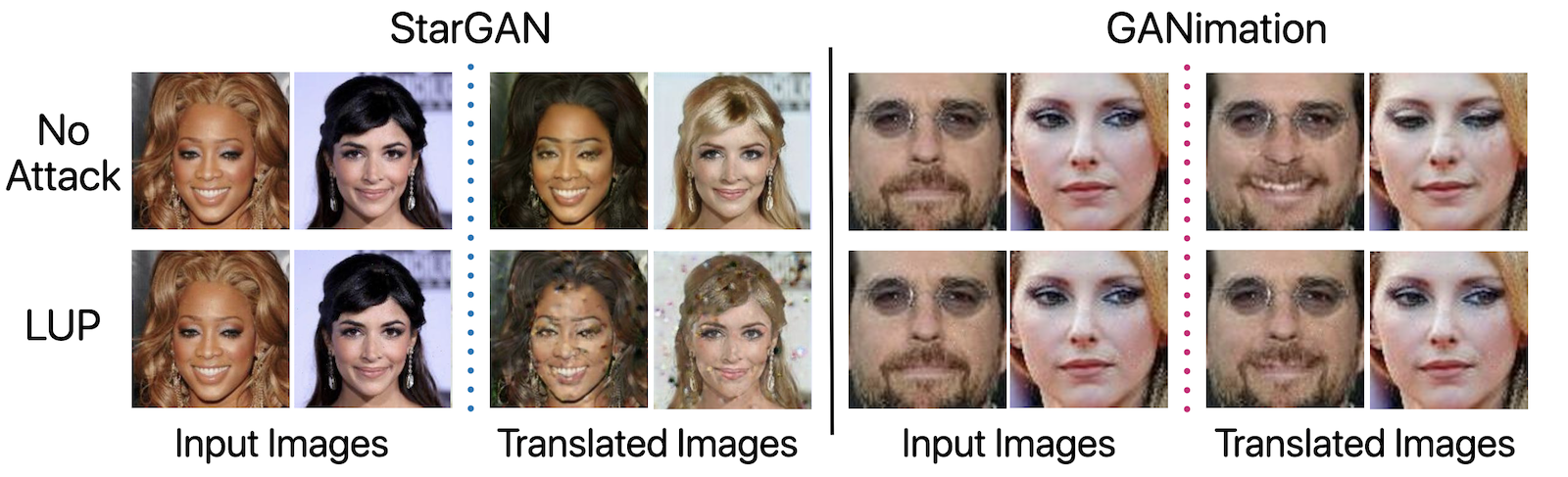}
    \vspace{-5pt}
    \caption{
    Qualitative examples of our LUP attack on StarGAN~\cite{choi2018stargan} and GANimation~\cite{pumarola2018ganimation}. We choose a \textit{maximum distortion attack} for StarGAN, where we successfully distort the output of the network and make it unusable. We chose an \textit{identity attack} for GANimation, where we successfully push the output of the network to be the same as the identity. Both approaches thwart deepfake generation.
    \label{fig:qualitative}}
    \vspace{-10pt}
\end{figure*}

\paragraph{Architectures and Datasets} 
We attack the GANimation~\cite{pumarola2018ganimation} and StarGAN~\cite{choi2018stargan} architectures. For GANimation we attack three expressions coded with distinct facial action units (AUs) and present average results. The expressions correspond to \textit{smile with closed eyes}, \textit{smile with open eyes} and \textit{eyebrow raise with smile}. These classes were selected because they enact very salient changes in the image, as opposed to other more subtle expressions. For StarGAN we present average results over 5 different attribute classes. The classes are \textit{black hair}, \textit{blond hair}, \textit{brown hair}, \textit{female} and \textit{old}. The dataset used for both architectures is the CelebA dataset~\cite{liu2015faceattributes}. For GANimation we attack 1,000 images using each expression, yielding 3,000 individual attacks. For StarGAN we attack 200 images using 5 different classes, yielding 1,000 individual attacks.

\paragraph{Implementation Details}
We use an adapted version of the NES and Bandits-TD implementations from the official Bandits-TD codebase. We modify the SimBA implementation from the official SimBA codebase. For NES and Bandits-TD we follow the parameter settings in the corresponding papers. For SimBA we use the same parameters used in the SimBA paper, except for the step size where we use 0.4 instead of 0.2. We use the same step size for LUP. We use a maximum number of saturating loss steps $n_\text{sat} = 20$ for LUP.

For the GANimation architecture, we build our leaked PCA components using 100 images, for each of the three expressions evaluated. We attack them using SimBA until we achieve the 0.005 success threshold or until we have performed 1,000 iterations of SimBA. We perform a total of 83,952 queries to construct these components (on average 27,984 queries for each expression). For StarGAN we build our PCA components using 10 images and 5 classes. We attack these images using SimBA until we achieve the 0.05 success threshold or until we have performed 1,000 iterations of SimBA. In total we perform 16,147 queries to build this basis.

\subsection{Experimental Results}

In this section we attack GANimation and StarGAN using IT-NES, IT-Bandits-TD, IT-SimBA and LUP. We compare the average number of queries required to successfully attack an image (by obtaining a loss under the designated threshold). We also present success rates and average perturbation magnitudes. 

\paragraph{GANimation}
We attack GANimation using an \textit{identity attack}, where we select the target image $\rTarget$ to be the input image, such that using our attack we push the network output to be the same as the input. We select a success threshold of $\threshold = 0.005$, meaning that we halt the attack when $L(\gen(\data + \perturb), \data) \leq \threshold$. After a successful attack at this threshold the transformations by GANimation are not noticeable. We use a maximum number of queries $\nqueries = 10,000$ for all methods. In Table \ref{table:attack_comparison} we show comparisons between LUP, IT-NES, IT-Bandits-TD and IT-SimBA. We can see that LUP is much more efficient than other methods achieving a \textasciitilde30\% reduction in average queries (551 vs. 393) compared to the next best method (IT-SimBA). Our method also achieves a lower average perturbation norm than the comparable IT-SimBA attack as well as an improved success rate. We present results for other thresholds in the supplementary material.

In Figure \ref{fig:ganimation_0005_hist} we present histograms of the number of queries needed to successfully attack images in the dataset. We can see that LUP is heavily skewed to the left and we achieve successful attacks using fewer than 20 queries for around half of the images in the dataset. This confirms our intuition that for certain architectures, the information collected during the \textit{leaking phase} allows us to build transferable attacks that resemble universal perturbations and that by exploiting this information we can construct effective attacks using very few queries for a large population of images. Figure \ref{fig:algos_cumulative_0005} shows the cumulative histogram of images successfully attacked for the number of queries represented by the x-axis. We observe that LUP achieves superior results than the competing methods. 

\paragraph{StarGAN}
We attack 200 images on the StarGAN architecture using 5 different attribute classes. We use a \textit{maximum distortion attack} (called \textit{optimal attack} in \cite{ruiz2020disrupting}), where the target image $\rTarget$ is the non-attacked output image $\gen(\data)$ and we maximize the loss instead of minimizing it to achieve the maximum amount of distortion in the output image. We present results for a threshold $\threshold = 0.05$, where the output image is visibly distorted. We use a maximum number of queries $\nqueries = 10,000$ for all methods. In Table \ref{table:attack_comparison} we show comparisons between IT-NES, IT-Bandits-TD, IT-SimBA and the \textit{exploitation phase} of LUP. We can see that LUP is much more efficient than other methods achieving a reduction in mean queries of \textasciitilde30\% compared to the next best attack and achieving a 100\% success rate. In this case the average norm is higher than competing methods. Qualitative analysis of images shows that the attack remains imperceptible (Figure \ref{fig:qualitative}). Additional qualitative examples and experiments with different thresholds are included in supplementary material.

In Figure \ref{fig:stargan_0050_hist} we show histograms of the number of queries required to attack images in the dataset. We can see that again LUP is heavily skewed to the left compared to competing methods. Figure \ref{fig:algos_cumulative_0005} shows the cumulative histogram of images successfully attacked for a specific number of queries.

One remark is that Bandits-TD, a method that achieves very good results in the image classification scenario on multiple networks, does not perform very well in this scenario. We believe that this is due to the time dependent prior. We observe that StarGAN presents strong random-seeming changes in the image output when a white-box attack of varying magnitude is applied. This seems to indicate that the gradient steps across time steps are not very correlated. We see the loss for Bandits-TD oscillate instead of strictly increase, which would seem to confirm our hypothesis.

\paragraph{Discussion}

The results demonstrate that LUP is more efficient than IT-SimBA during the exploitation phase. This is a consequence of the transferability of the leaked PCA components that are subsequently used as candidate vectors during the exploitation phase. We find that image translation architectures have specific vulnerabilities and that there exist correlations between attacks constructed for different images. This is the nugget of intuition that motivates our proposed approach.

By extrapolation, we estimate that the total amount of queries to attack 100,000 images for the GANimation architecture using the LUP algorithm is 39,383,952 - counting both the leaking and exploitation phases. Using IT-SimBA the total amount of queries would be 55,100,000 - \textbf{roughly 15.7 million more queries}. When scaling to a real service, which protects images of individuals from deepfake generation, that service will be expected to protect millions of images. We conclude that this reduction in the total number of black-box queries can save significant resources.

\section{Conclusion}

In this work we present the first successful black-box attacks on image translation systems, with an application to disrupting the generation of deepfake images. This is a step forward in the war on deepfakes, since many modern deepfake generation systems can manipulate a face using only one image and no longer need to rely on face swapping or large datasets of images of a person. Our work can be directly used to protect photographs of people from being modified by deep learning facial manipulation systems in the real world by attacking these systems.

A key limitation of many existing black-box attacks on classifiers, that we modify for applicability to the image translation scenario, is the high number of queries needed to generate adversarial attacks. We present a simple, yet highly effective method, called Leaking Universal Perturbations (LUP), that reduces the total number of queries necessary to find attacks for a dataset of images by extracting information from initial attacks during a leaking phase. This information is then used during an exploitation phase to achieve more efficient attacks. We demonstrate the effectiveness of our method on both the GANimation and StarGAN architectures on the CelebA dataset.

\section*{Broader Impact} 

Deepfake generation technology has been advancing at an ever accelerating pace and has already achieved impressive results. There are several compelling potential applications of the technology, such as face frontalization, actor editing in movies, correcting expressions, pose or lighting in pictures, and realistic avatar generation. We believe that many of these applications are either neutral or good for society as a whole. Unfortunately, deepfakes have several potential applications that are incontestably negative and morally wrong. In this work we have discussed some of these applications, that have already materialized, such as non-consensual pornographic footage generation and fake political speeches.

We firmly believe that a person has an inalienable right to protect their likeness. Our work is a first step in this direction. The ultimate goal of our work is to enable a person to have the choice of giving or withholding consent when it comes to deepfake generation of their likeness. With advances in black-box attacks on generative models, we believe there is a chance of achieving this goal. If a security arms race were to begin in this area, we believe that the mere existence of protection mechanisms for people's images will increase the cost of generating non-consensual deepfakes.

\bibliographystyle{abbrv}
\bibliography{egbib}

\begin{thebibliography}{10}

\bibitem{bashkirova2019adversarial}
D.~Bashkirova, B.~Usman, and K.~Saenko.
\newblock Adversarial self-defense for cycle-consistent gans.
\newblock In {\em Advances in Neural Information Processing Systems}, pages
  635--645, 2019.

\bibitem{carlini2017towards}
N.~Carlini and D.~Wagner.
\newblock Towards evaluating the robustness of neural networks.
\newblock In {\em 2017 IEEE Symposium on Security and Privacy (SP)}, pages
  39--57. IEEE, 2017.

\bibitem{chen2017zoo}
P.-Y. Chen, H.~Zhang, Y.~Sharma, J.~Yi, and C.-J. Hsieh.
\newblock Zoo: Zeroth order optimization based black-box attacks to deep neural
  networks without training substitute models.
\newblock In {\em Proceedings of the 10th ACM Workshop on Artificial
  Intelligence and Security}, pages 15--26, 2017.

\bibitem{Cheng2019QueryEfficientHB}
M.~Cheng, T.~Le, P.~Chen, H.~Zhang, J.~Yi, and C.~Hsieh.
\newblock Query-efficient hard-label black-box attack: An optimization-based
  approach.
\newblock In {\em 7th International Conference on Learning Representations,
  {ICLR} 2019, New Orleans, LA, USA, May 6-9, 2019}. OpenReview.net, 2019.

\bibitem{choi2018stargan}
Y.~Choi, M.~Choi, M.~Kim, J.-W. Ha, S.~Kim, and J.~Choo.
\newblock Stargan: Unified generative adversarial networks for multi-domain
  image-to-image translation.
\newblock In {\em Proceedings of the IEEE conference on computer vision and
  pattern recognition}, pages 8789--8797, 2018.

\bibitem{choi2019stargan}
Y.~Choi, Y.~Uh, J.~Yoo, and J.-W. Ha.
\newblock Stargan v2: Diverse image synthesis for multiple domains.
\newblock {\em arXiv preprint arXiv:1912.01865}, 2019.

\bibitem{chu2017cyclegan}
C.~Chu, A.~Zhmoginov, and M.~Sandler.
\newblock Cyclegan, a master of steganography.
\newblock {\em arXiv preprint arXiv:1712.02950}, 2017.

\bibitem{ekman1997face}
R.~Ekman.
\newblock {\em What the face reveals: Basic and applied studies of spontaneous
  expression using the Facial Action Coding System (FACS)}.
\newblock Oxford University Press, USA, 1997.

\bibitem{gandhi2020adversarial}
A.~Gandhi and S.~Jain.
\newblock Adversarial perturbations fool deepfake detectors.
\newblock {\em arXiv preprint arXiv:2003.10596}, 2020.

\bibitem{explaining_adv}
I.~Goodfellow, J.~Shlens, and C.~Szegedy.
\newblock Explaining and harnessing adversarial examples.
\newblock In {\em Proc. ICLR}, 2015.

\bibitem{guera2018deepfake}
D.~G{\"u}era and E.~J. Delp.
\newblock Deepfake video detection using recurrent neural networks.
\newblock In {\em 2018 15th IEEE International Conference on Advanced Video and
  Signal Based Surveillance (AVSS)}, pages 1--6. IEEE, 2018.

\bibitem{guo2019simple}
C.~Guo, J.~R. Gardner, Y.~You, A.~G. Wilson, and K.~Q. Weinberger.
\newblock Simple black-box adversarial attacks.
\newblock In K.~Chaudhuri and R.~Salakhutdinov, editors, {\em Proceedings of
  the 36th International Conference on Machine Learning, {ICML} 2019, 9-15 June
  2019, Long Beach, California, {USA}}, volume~97 of {\em Proceedings of
  Machine Learning Research}, pages 2484--2493. {PMLR}, 2019.

\bibitem{ilyas2018black}
A.~Ilyas, L.~Engstrom, A.~Athalye, and J.~Lin.
\newblock Black-box adversarial attacks with limited queries and information.
\newblock In {\em Proceedings of the 35th International Conference on Machine
  Learning, {ICML} 2018, Stockholmsm{\"{a}}ssan, Stockholm, Sweden, July 10-15,
  2018}, pages 2142--2151, 2018.

\bibitem{IEM2018PriorCB}
A.~Ilyas, L.~Engstrom, and A.~Madry.
\newblock Prior convictions: Black-box adversarial attacks with bandits and
  priors.
\newblock In {\em 7th International Conference on Learning Representations,
  {ICLR} 2019, New Orleans, LA, USA, May 6-9, 2019}. OpenReview.net, 2019.

\bibitem{isola2017image}
P.~Isola, J.-Y. Zhu, T.~Zhou, and A.~A. Efros.
\newblock Image-to-image translation with conditional adversarial networks.
\newblock In {\em Proceedings of the IEEE conference on computer vision and
  pattern recognition}, pages 1125--1134, 2017.

\bibitem{kim2018deep}
H.~Kim, P.~Garrido, A.~Tewari, W.~Xu, J.~Thies, M.~Nie{\ss}ner, P.~P{\'e}rez,
  C.~Richardt, M.~Zollh{\"o}fer, and C.~Theobalt.
\newblock Deep video portraits.
\newblock {\em ACM Transactions on Graphics (TOG)}, 37(4):1--14, 2018.

\bibitem{kos2018adversarial}
J.~Kos, I.~Fischer, and D.~Song.
\newblock Adversarial examples for generative models.
\newblock In {\em 2018 IEEE Security and Privacy Workshops (SPW)}, pages
  36--42. IEEE, 2018.

\bibitem{kurakin2016adversarial}
A.~Kurakin, I.~J. Goodfellow, and S.~Bengio.
\newblock Adversarial examples in the physical world.
\newblock In {\em 5th International Conference on Learning Representations,
  {ICLR} 2017, Toulon, France, April 24-26, 2017, Workshop Track Proceedings}.
  OpenReview.net, 2017.

\bibitem{li2019exposing}
Y.~Li and S.~Lyu.
\newblock Exposing deepfake videos by detecting face warping artifacts.
\newblock In {\em Proceedings of the IEEE Conference on Computer Vision and
  Pattern Recognition Workshops}, pages 46--52, 2019.

\bibitem{liu2016delving}
Y.~Liu, X.~Chen, C.~Liu, and D.~Song.
\newblock Delving into transferable adversarial examples and black-box attacks.
\newblock {\em CoRR}, abs/1611.02770, 2016.

\bibitem{liu2015faceattributes}
Z.~Liu, P.~Luo, X.~Wang, and X.~Tang.
\newblock Deep learning face attributes in the wild.
\newblock In {\em Proceedings of International Conference on Computer Vision
  (ICCV)}, December 2015.

\bibitem{madry2018towards}
A.~Madry, A.~Makelov, L.~Schmidt, D.~Tsipras, and A.~Vladu.
\newblock Towards deep learning models resistant to adversarial attacks.
\newblock In {\em International Conference on Learning Representations}, 2018.

\bibitem{moosavi2017universal}
S.-M. Moosavi-Dezfooli, A.~Fawzi, O.~Fawzi, and P.~Frossard.
\newblock Universal adversarial perturbations.
\newblock In {\em Proceedings of the IEEE conference on computer vision and
  pattern recognition}, pages 1765--1773, 2017.

\bibitem{moosavi2016deepfool}
S.-M. Moosavi-Dezfooli, A.~Fawzi, and P.~Frossard.
\newblock Deepfool: a simple and accurate method to fool deep neural networks.
\newblock In {\em Proceedings of the IEEE conference on computer vision and
  pattern recognition}, pages 2574--2582, 2016.

\bibitem{narodytska2016simple}
N.~Narodytska and S.~P. Kasiviswanathan.
\newblock Simple black-box adversarial perturbations for deep networks.
\newblock {\em arXiv preprint arXiv:1612.06299}, 2016.

\bibitem{neekhara2020adversarial}
P.~Neekhara, S.~Hussain, M.~Jere, F.~Koushanfar, and J.~McAuley.
\newblock Adversarial deepfakes: Evaluating vulnerability of deepfake detectors
  to adversarial examples.
\newblock {\em arXiv preprint arXiv:2002.12749}, 2020.

\bibitem{nguyen2015deep}
A.~Nguyen, J.~Yosinski, and J.~Clune.
\newblock Deep neural networks are easily fooled: High confidence predictions
  for unrecognizable images.
\newblock In {\em Proceedings of the IEEE conference on computer vision and
  pattern recognition}, pages 427--436, 2015.

\bibitem{papernot2017practical}
N.~Papernot, P.~McDaniel, I.~Goodfellow, S.~Jha, Z.~B. Celik, and A.~Swami.
\newblock Practical black-box attacks against machine learning.
\newblock In {\em Proceedings of the 2017 ACM on Asia conference on computer
  and communications security}, pages 506--519. ACM, 2017.

\bibitem{papernot2016limitations}
N.~Papernot, P.~McDaniel, S.~Jha, M.~Fredrikson, Z.~B. Celik, and A.~Swami.
\newblock The limitations of deep learning in adversarial settings.
\newblock In {\em 2016 IEEE European Symposium on Security and Privacy
  (EuroS\&P)}, pages 372--387. IEEE, 2016.

\bibitem{pumarola2018ganimation}
A.~Pumarola, A.~Agudo, A.~M. Martinez, A.~Sanfeliu, and F.~Moreno-Noguer.
\newblock Ganimation: Anatomically-aware facial animation from a single image.
\newblock In {\em Proceedings of the European Conference on Computer Vision
  (ECCV)}, pages 818--833, 2018.

\bibitem{Rossler_2019_ICCV}
A.~Rossler, D.~Cozzolino, L.~Verdoliva, C.~Riess, J.~Thies, and M.~Niessner.
\newblock Faceforensics++: Learning to detect manipulated facial images.
\newblock In {\em The IEEE International Conference on Computer Vision (ICCV)},
  October 2019.

\bibitem{ruiz2020disrupting}
N.~Ruiz, S.~A. Bargal, and S.~Sclaroff.
\newblock Disrupting deepfakes: Adversarial attacks against conditional image
  translation networks and facial manipulation systems.
\newblock {\em CoRR}, abs/2003.01279, 2020.

\bibitem{szegedy2013intriguing}
C.~Szegedy, W.~Zaremba, I.~Sutskever, J.~Bruna, D.~Erhan, I.~Goodfellow, and
  R.~Fergus.
\newblock Intriguing properties of neural networks.
\newblock In {\em In Proc. ICLR}, 2014.

\bibitem{tabacof2016adversarial}
P.~Tabacof, J.~Tavares, and E.~Valle.
\newblock Adversarial images for variational autoencoders.
\newblock {\em arXiv preprint arXiv:1612.00155}, 2016.

\bibitem{tewari2020stylerig}
A.~Tewari, M.~Elgharib, G.~Bharaj, F.~Bernard, H.-P. Seidel, P.~P{\'e}rez,
  M.~Zollh{\"o}fer, and C.~Theobalt.
\newblock Stylerig: Rigging stylegan for 3d control over portrait images.
\newblock {\em arXiv preprint arXiv:2004.00121}, 2020.

\bibitem{thies2016face2face}
J.~Thies, M.~Zollhofer, M.~Stamminger, C.~Theobalt, and M.~Nie{\ss}ner.
\newblock Face2face: Real-time face capture and reenactment of rgb videos.
\newblock In {\em Proceedings of the IEEE conference on computer vision and
  pattern recognition}, pages 2387--2395, 2016.

\bibitem{thies2018headon}
J.~Thies, M.~Zollh{\"o}fer, C.~Theobalt, M.~Stamminger, and M.~Nie{\ss}ner.
\newblock Headon: Real-time reenactment of human portrait videos.
\newblock {\em ACM Transactions on Graphics (TOG)}, 37(4):1--13, 2018.

\bibitem{tu2019autozoom}
C.-C. Tu, P.~Ting, P.-Y. Chen, S.~Liu, H.~Zhang, J.~Yi, C.-J. Hsieh, and S.-M.
  Cheng.
\newblock Autozoom: Autoencoder-based zeroth order optimization method for
  attacking black-box neural networks.
\newblock In {\em Proceedings of the AAAI Conference on Artificial
  Intelligence}, volume~33, pages 742--749, 2019.

\bibitem{Usman_2019_ICCV}
B.~Usman, N.~Dufour, K.~Saenko, and C.~Bregler.
\newblock Puppetgan: Cross-domain image manipulation by demonstration.
\newblock In {\em The IEEE International Conference on Computer Vision (ICCV)},
  October 2019.

\bibitem{wangfakespotter}
R.~Wang, L.~Ma, F.~Juefei-Xu, X.~Xie, J.~Wang, and Y.~Liu.
\newblock Fakespotter: A simple baseline for spotting ai-synthesized fake
  faces.
\newblock {\em arXiv preprint arXiv:1909.06122}, 2019.

\bibitem{wang2019cnngenerated}
S.-Y. Wang, O.~Wang, R.~Zhang, A.~Owens, and A.~A. Efros.
\newblock Cnn-generated images are surprisingly easy to spot...for now.
\newblock In {\em CVPR}, 2020.

\bibitem{wang2018high}
T.-C. Wang, M.-Y. Liu, J.-Y. Zhu, A.~Tao, J.~Kautz, and B.~Catanzaro.
\newblock High-resolution image synthesis and semantic manipulation with
  conditional gans.
\newblock In {\em Proceedings of the IEEE conference on computer vision and
  pattern recognition}, pages 8798--8807, 2018.

\bibitem{8683164}
X.~{Yang}, Y.~{Li}, and S.~{Lyu}.
\newblock Exposing deep fakes using inconsistent head poses.
\newblock In {\em ICASSP 2019 - 2019 IEEE International Conference on
  Acoustics, Speech and Signal Processing (ICASSP)}, pages 8261--8265, 2019.

\bibitem{zakharov2019few}
E.~Zakharov, A.~Shysheya, E.~Burkov, and V.~Lempitsky.
\newblock Few-shot adversarial learning of realistic neural talking head
  models.
\newblock In {\em Proceedings of the IEEE International Conference on Computer
  Vision}, pages 9459--9468, 2019.

\bibitem{zhu2017unpaired}
J.-Y. Zhu, T.~Park, P.~Isola, and A.~A. Efros.
\newblock Unpaired image-to-image translation using cycle-consistent
  adversarial networks.
\newblock In {\em Proceedings of the IEEE international conference on computer
  vision}, pages 2223--2232, 2017.

\end{thebibliography}

\clearpage

\begin{center}  
 \begin{huge}  
 \textbf{Supplementary Material}
\end{huge}  
\end{center}

\ \\

\section*{Experiments With Varying Thresholds}

In this section we present comparisons of attacks on StarGAN and GANimation with different thresholds. In the main manuscript we presented results for GANimation under the threshold $\threshold = 0.005$. We now present results for threshold $\threshold = 0.01$ in Table \ref{table:ganimation_supp1}. Similarly, the main manuscript presented results for StarGAN under the threshold $\threshold = 0.005$. Table \ref{table:stargan_supp1} show results for StarGAN under the threshold $\threshold = 0.025$ for a maximum distortion attack. We observe that LUP achieves successful attacks using substantially lower number of average queries compared to IT-NES, IT-Bandits-TD and IT-SimBA.

\begin{figure*}[t]
    \centering
    \includegraphics[width=0.48\textwidth]{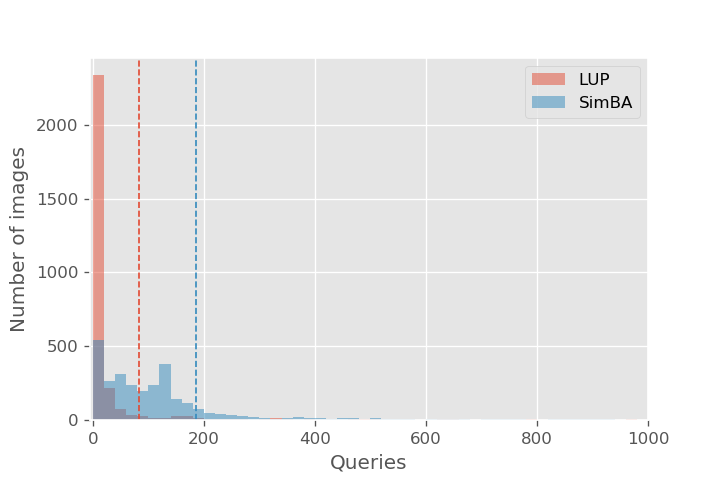}
    \includegraphics[width=0.48\textwidth]{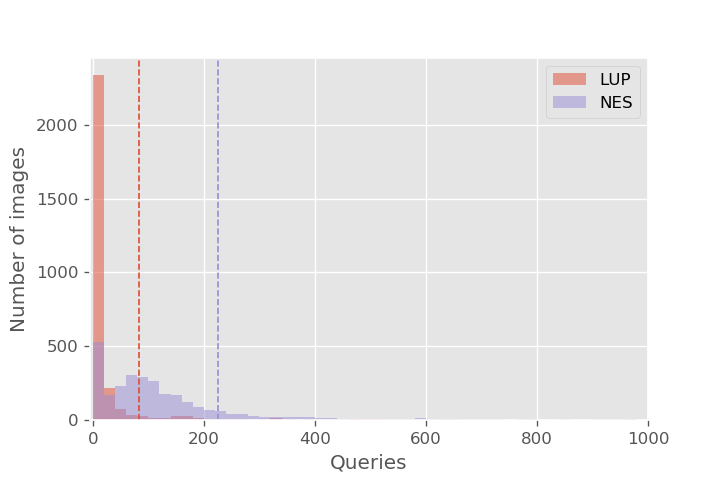}
    \caption{
    Histogram of queries required for successful attacks on GANimation (3,000 attacks, three expressions), using success threshold $\threshold=0.01$. Vertical lines show mean queries for each method.
    \label{fig:ganimation_0010_hist}}
\end{figure*}

\begin{figure*}[t]
    \centering
    \includegraphics[width=0.48\textwidth]{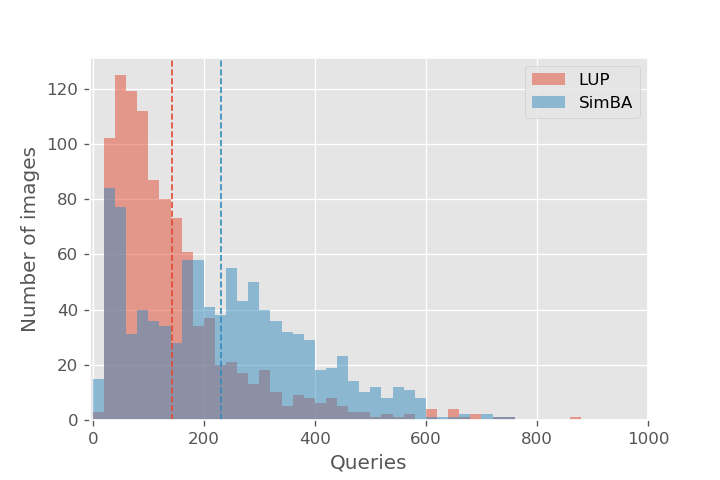}
    \includegraphics[width=0.48\textwidth]{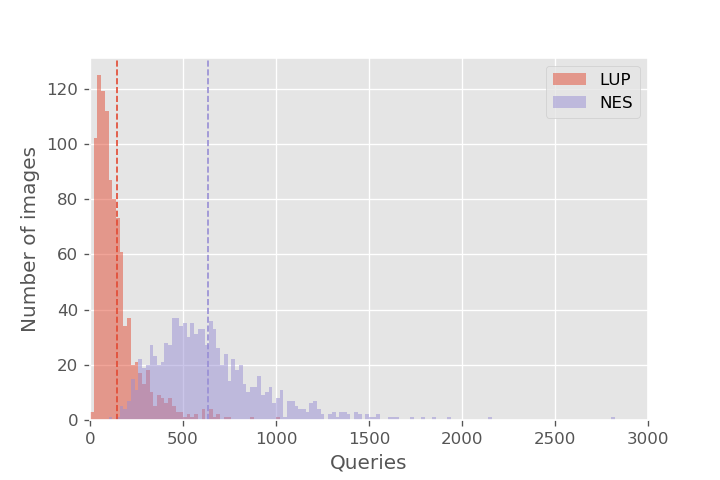}
    \caption{Histogram of queries required for a successful attack on StarGAN of 1,000 attacks (five classes). We use a success threshold $\threshold=0.025$. Vertical lines represent mean queries for each method.
    \label{fig:stargan_0025_hist}}
\end{figure*}

We also show the corresponding histograms of successful attacks for LUP, IT-SimBA and IT-NES for GANimation in Figure \ref{fig:ganimation_0010_hist} ($\threshold = 0.01$) and StarGAN in Figure \ref{fig:stargan_0025_hist} ($\threshold = 0.025$). For both networks we observe distributions for LUP that are very skewed to the left, compared to IT-SimBA and IT-NES. In particular LUP is able to successfully complete more than 2,500 out of 3,000 attacks using 20 or less queries for GANimation using threshold $\threshold = 0.01$ (Figure \ref{fig:ganimation_0010_hist}).

\begin{figure}[t]
\includegraphics[clip,width=0.48\columnwidth]{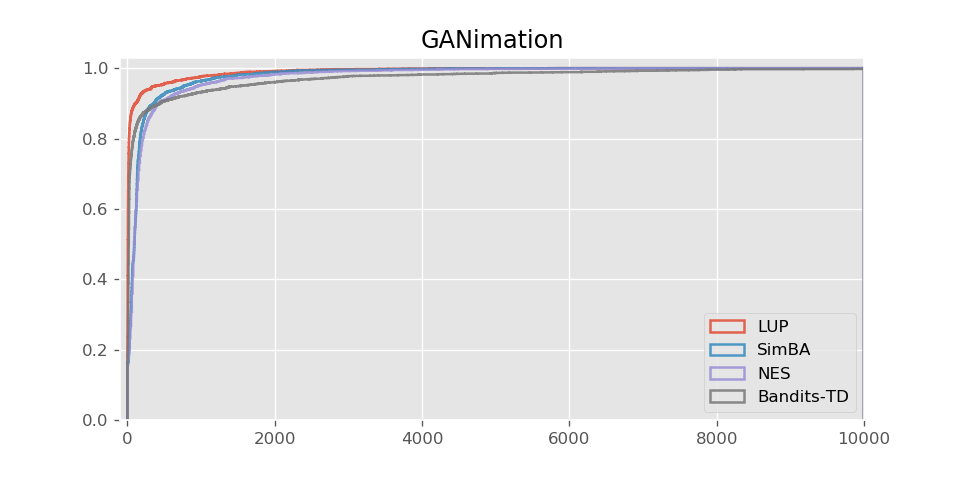}
\includegraphics[clip,width=0.48\columnwidth]{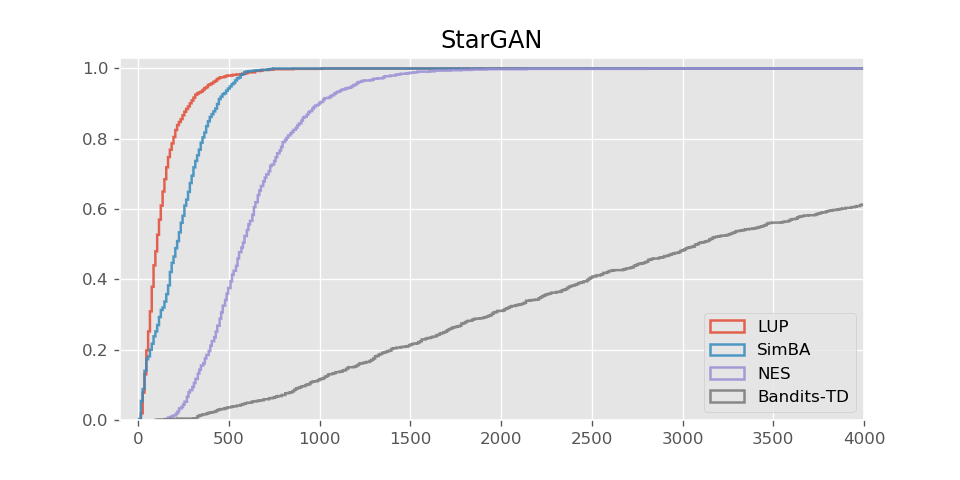}
\caption[]{Success rate by number of queries on 3,000 attacks on GANimation for three expressions (left) and 1,000 attacks on StarGAN for five attribute classes (right). We use success threshold $\threshold=0.01$ for GANimation and threshold $\threshold=0.025$ for StarGAN. Vertical lines represent mean queries for each method.
\label{fig:algos_cumulative_supp}}
\end{figure}

Figure \ref{fig:algos_cumulative_supp} shows the success rate curves for GANimation and StarGAN under the thresholds $\threshold = 0.01$, and $\threshold = 0.025$, respectively. We retrieve the same result patterns as the main manuscript, with LUP performing best among competing algorithms.

\section*{Additional Qualitative Samples}

\begin{figure*}[t]
    \centering
    \includegraphics[width=1\textwidth]{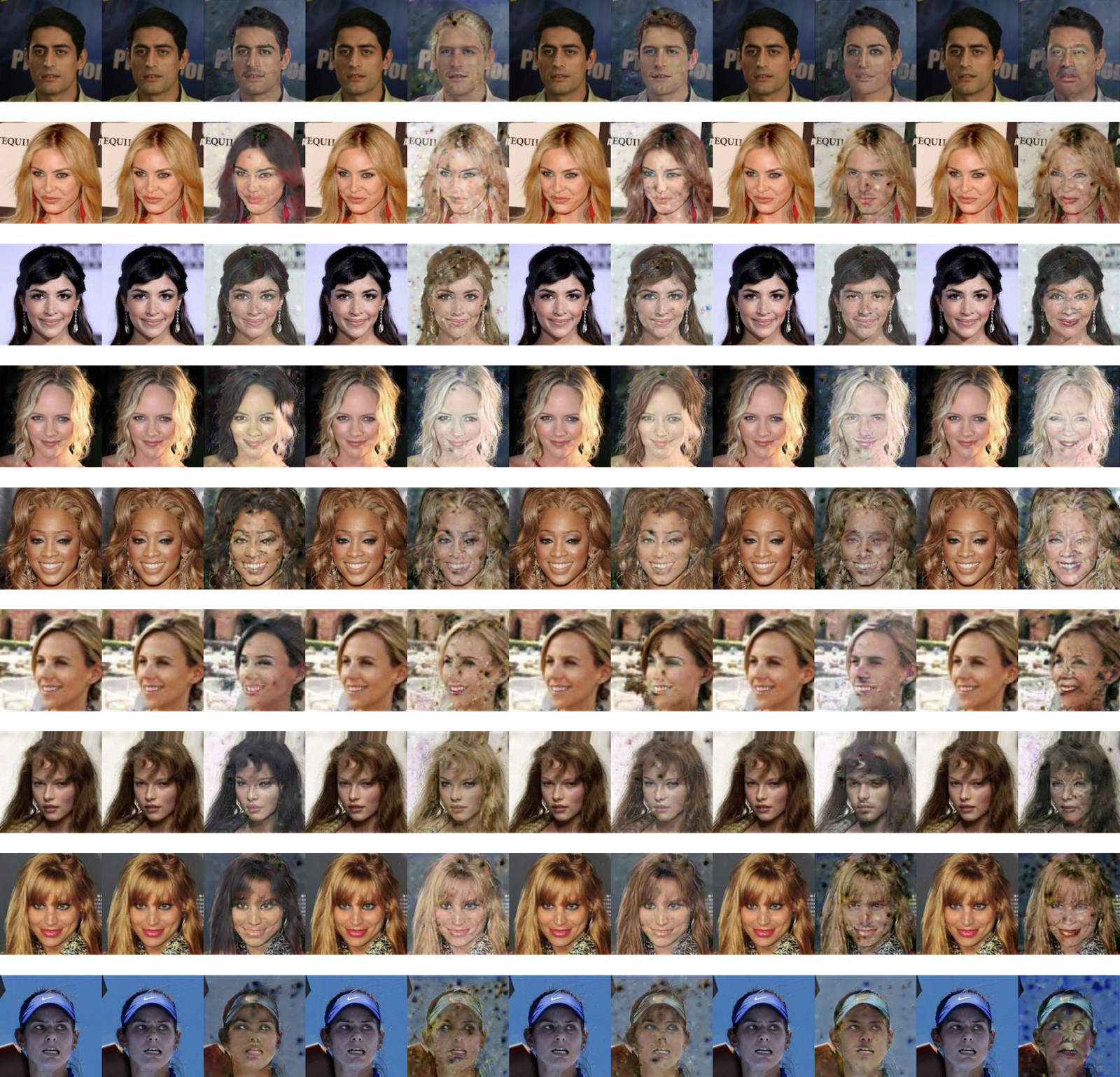}
    \caption{
    Qualitative examples of our LUP attack on StarGAN~\cite{choi2018stargan}. Using a \textit{maximum distortion attack} where we successfully distort the output of the network and make it unusable. We show results for five different attribute classes \textit{black hair}, \textit{blond hair}, \textit{brown hair}, \textit{female} and \textit{old}, in this order. We first show the unaltered, original input image in the first column. Then, for each attribute we show (1) the attacked input image (2) the output of StarGAN using the attacked input, for this attribute modification. We show these successively for each attribute.
    \label{fig:stargan_supp_samples}}
\end{figure*}

We show samples of attacked images for StarGAN using LUP in Figure \ref{fig:stargan_supp_samples}. We present the original image, the attacked input images and the disrupted output images for different samples and for all five classes used in the main manuscript (\textit{black hair}, \textit{blond hair}, \textit{brown hair}, \textit{female} and \textit{old}). We observe that all of the translated images have been disrupted and can no longer be used for their original purpose.

\begin{figure*}[t]
    \centering
    \includegraphics[width=1\textwidth]{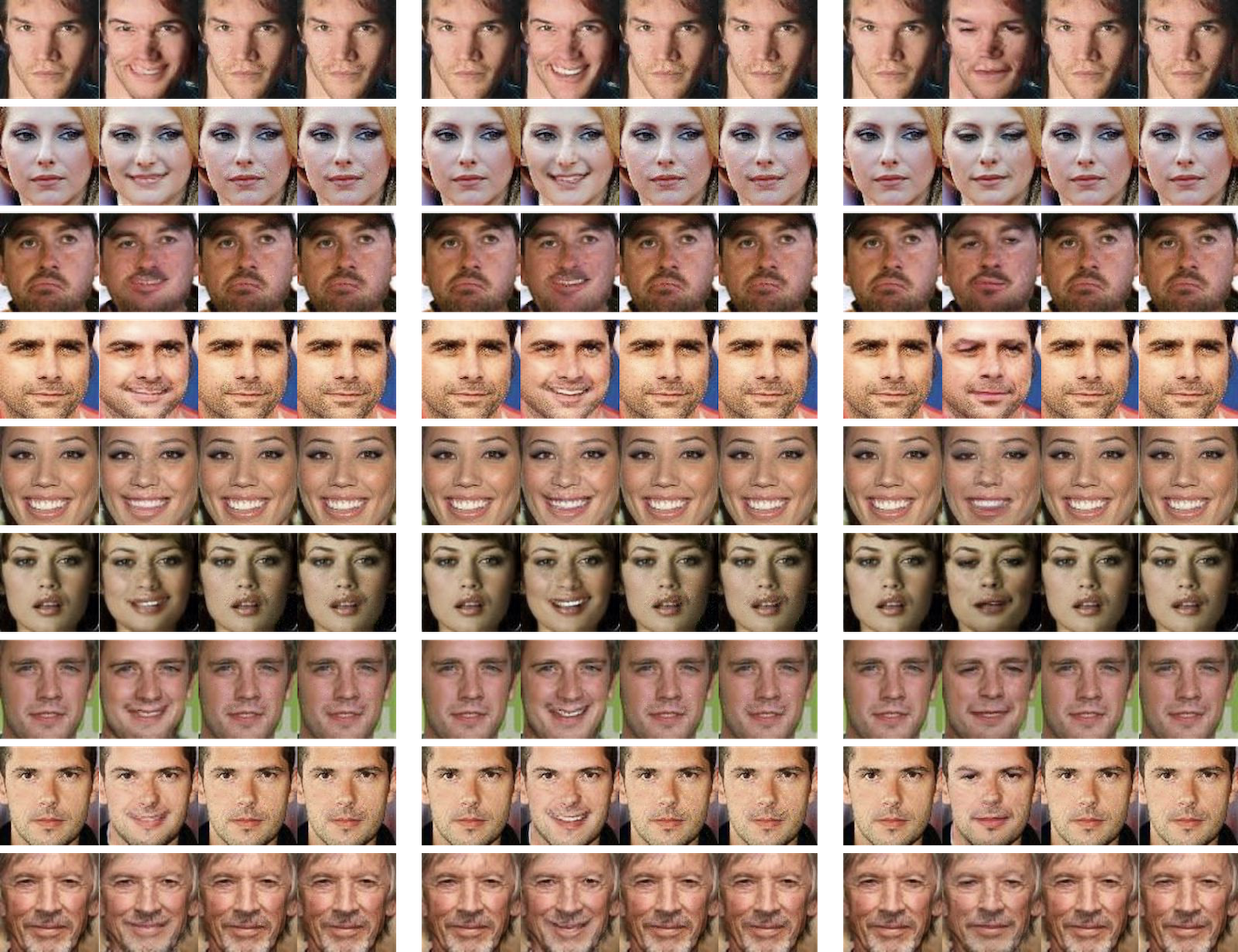}
    \caption{
    Qualitative examples of our LUP attack on GANimation~\cite{pumarola2018ganimation}. Using an \textit{identity attack} where we successfully push the output of the network to be the same as the identity. We show the attack for three different expressions. \textit{Eyebrow raise with smile} (left), \textit{smile with open eyes} (middle) and \textit{smile with closed eyes} (right). For each expression column we show (1) the original input image (2) the ground-truth, non-attacked output of GANimation (3) the attacked input (4) the output of GANimation using the attacked input.
    \label{fig:ganimation_supp_samples}}
\end{figure*}

We show samples of attacked images for GANimation using LUP in Figure \ref{fig:ganimation_supp_samples}. We use an identity attack, where we push the output of the attacked image to be the same as the input image (i.e. we disable modification by GANimation). We test on three different expressions, corresponding to \textit{eyebrow raise with smile}, \textit{smile with closed eyes} and \textit{smile with open eyes}. We observe that all of the attacked outputs are very similar to the input and there are no perceptible modifications by the GANimation network. We can also observe that the attacks on the inputs are imperceptible and thus do not lower the quality of the input image.

\section*{Details on GANimation and StarGAN}

\paragraph{GANimation} 

We use an open-source implementation of GANimation trained for 37 epochs on the CelebA~\cite{liu2015faceattributes} dataset for 17 action units (AU) from the Facial Action Unit Coding System (FACS)~\cite{ekman1997face}. GANimation takes as input a vector of facial Action Units (AUs). For the entirety of our work we selected three sufficiently distinct AU vectors, each describing a different expression. The three AU vectors that we selected correspond to: \textit{eyebrow raise with smile}, \textit{smile with closed eyes}, \textit{smile with open eyes}.

\paragraph{StarGAN} 

We use the official open-source implementation of StarGAN, trained on the CelebA dataset~\cite{liu2015faceattributes} for the five attributes \textit{black hair}, \textit{blond hair}, \textit{brown hair}, \textit{female} and \textit{old}. We use these same attribute classes for our attacks.

\begin{table}[t]
\centering
  \caption{Attack comparison of \textit{identity attack} on GANimation with threshold $\threshold = 0.01$ using 3 expressions. We show the mean number of queries per image, as well as the average norm of the perturbation and the success rate percentage.
  \label{table:ganimation_supp1}}
\resizebox{0.70\textwidth}{!}{
  \begin{tabular}{cccc}
    \toprule
    Attack & Avg. Queries & Avg. Norm & Success Rate \\
    \midrule
    IT-NES & 226 & 1.05 & 100\% \\
    IT-Bandits-TD & 254 & 3.20 & 99.8\%  \\
    IT-SimBA & 186 & 2.97 & 100\% \\
    LUP & \textbf{83} & 1.49 & 100\% \\
    \bottomrule
  \end{tabular}
}
\end{table}

\begin{table}[t]
\centering
  \caption{Attack comparison of \textit{maximum distortion attacks} on StarGAN for threshold $\threshold = 0.025$ using 5 attributes. We show the mean number of queries per image, as well as the average norm of the perturbation and the success rate percentage.
  \label{table:stargan_supp1}}
\resizebox{0.70\textwidth}{!}{
  \begin{tabular}{cccc}
    \toprule
    Attack & Avg. Queries & Avg. Norm & Success Rate \\
    \midrule
    IT-NES & 633 & 2.32 & 100\% \\
    IT-Bandits-TD & 3229 & 5.00 & 86.8\%  \\
    IT-SimBA & 232 & 4.41 & 100\% \\
    LUP & \textbf{143} & 4.26 & 100\% \\
    \bottomrule
  \end{tabular}
}
\end{table}

\end{document}